\documentclass[conference]{IEEEtran}
\IEEEoverridecommandlockouts
\usepackage{cite}
\usepackage{amsmath,amssymb,amsfonts}
\usepackage{algorithmic}
\usepackage{graphicx}
\usepackage{textcomp}
\usepackage{xcolor}
\usepackage{url}
\def\BibTeX{{\rm B\kern-.05em{\sc i\kern-.025em b}\kern-.08em
    T\kern-.1667em\lower.7ex\hbox{E}\kern-.125emX}}

\begin{document}
	
\urlstyle{same}

\title{Multi-area Target Individual Detection with Free Drawing on Video\\
{\footnotesize \textsuperscript{*}}
\thanks{}
}

\author{\IEEEauthorblockN{1\textsuperscript{st} Jinwei Lin}
\IEEEauthorblockA{\textit{Monash University} \\
\textit{School of Information Technology}\\
Selangor, Malaysia \\
0000-0003-0558-6699}
}

\maketitle

\begin{abstract}
This paper has provided a novel design idea and some implementation methods to make a real time detection of multi-areas with multiple detecting areas that are generated by the real time drawing on the screen display of the video. The drawing on the video will remain the output as polylines, and the colors of the outlines will change when the stage of drawing or detecting is changed. The shape of the drawn area is free to be customized and real-time effective. The configuration of the drawn areas can be renewed and the detecting areas are working individually. The detection result should be shown with a GUI designed by Tkinter. The object recognition model was developed on YOLOv5 but can be changed to others, which means the core design and implementation idea of this paper is model-independent. With PIL and OpenCV and Tkinter, the drawing effect is real time and efficient. The design and code of this research is basic and can be extended to be implemented in numerous monitoring and detecting situations. 
\end{abstract}

\section{Introduction}
The research of people detection or objects detection is considered as one of the primary areas of the field of computer vision. The detection of objects, especially the detection of humans is significant in current technology implementations. In the field of detection or recognition of humans, the detection of the location and the target tracking of humans are meaningful for various application situations. For some detecting situations, one or multiple cameras are used to make a single detection for the human traffic or human flow distribution transform. When it comes to the implementation of that using one single camera to detect multiple areas, the normal methods to address this issue will prefer the design solution that tries to meet these implementation requirements on multiple cameras, instead of handling them on one camera scene. For the real-world applications, the cost of project deployment and resource consumption of the equipment should be considered as a key factor. 

If there is an applicable solution or method that can detect multiple areas of one visual area with one camera, it will save most of the computed resource consumption and hardwares arrangement cost,compared with the scheme that uses multiple cameras to detect multiple areas, which is the core design and research topic of this paper. Beside the decreasing of the consumption of hardwares layout and computed resources, the multiple target individual detection method of this paper can intelligently set and assign the current whole visual area into multiple sub-area of detection in real-time, which means that the management of  arrangement of multiple detecting areas will be easy and free. With the easy-configured function and APIs, various of the corresponding implementations can be developed based on the flexible setting and designing. For instance, detecting the traffic flow transform of multiple areas on one specific area with one single camera, and detecting and judging whether the location of multiple objects is right or wrong. Moreover, detecting the important sub-areas of the visual scope of security and protection systems. The code source of the research of this paper is open source on GitHub: \url{https://github.com/JYLinOK/FreedrawingDetecting}.


\section{Literature Review}
\label{sec:literature}
\subsection{Drawing on Video}
When using a video window to display the real-time or historical scene captured by the cameras, it will be useful and convenient to make the marks or key notes or individual managements on the scene by finding a method that can draw the lines or polylines that present the special areas. Drawing on video can make the video be more interactive and functional, compared with the pure videos. There is not much research on how to draw the single polylines on a real-time video image. However, this kind of research is useful when it comes to some special applications. Computer drawing has a wide range of applications \cite{Freeman1974Computer}. As a common drawing toolkit or control, canvas is designed as a basic control or component of the function model of drawing. From the fundamental analysis, the video consists of a series of rapidly switching images in a continuous time. Therefore, one approach to drawing the polylines on the video is to draw the polylines on the individual images of the video. This process should be handled in real-time to make the low latency and high efficient edited video.

\subsection{Drawing on Video}
Object detection is considered as one significant research area of computer vision \cite{1998General}, which means using the computer algorithms to deal with the detection tasks or issues. There is numerous related research about this scope \cite{2018Advanced}. Object detection, especially the detection of human objects, is applied in various real practical implementations, which causes the rapid development of this research field. As the technologies develop, there are various excellent processing algorithms, of which the YOLOv5 is one representative deep learning  model. The YOLOv5 model is open source in \url{https://github.com/ultralytics/yolov5}. YOLOv5 can be used to detect the objects of multiple classes and have a peer leading process speed on object detection. The processing algorithms of this paper will be developed with the interface section and data processing of YOLOv5 but not limited, which means other excellent detection models are also be recommended and can be connected or combined with the core algorithms of this paper.

\subsection{Tkinter and Python GUI}
Python, which is widely used in data analysis and scientific research fields, is accepted as the programming language of algorithm implementations of this paper. As three of the main computer vision libraries of python, OpenCV, PIL and Tkinter are used to make the UI (i.e.: User Interface) implementation of this paper. Using Tkinter, the native library of Python, some simple and fundamental GUI (i.e.: Graphical User Interface) projects can be easily programmed \cite{2020Design}. The first step to draw on a video screen real-time display is to draw images of the video on the user interface window, which can be implemented by the component canvas of Tkinter. Subsequently, the approach to draw the polylines on the canvas or image should be considered. Using the libraries Numpy and PIL can make it able to transform the image data between Numpy arrays and PIL image format objects, which is useful to draw the graphic information on the video image and make relative processing. It is difficult to configure the transparency of the canvas component of Tkinter while remaining the drawing information on the canvas. Therefore, directly drawing the graphic information on the video image is the primary design principle and implementation method of this paper.

\subsection{Multi-area Objects Detection}

Multi-area objects detection is the detection that uses one or multiple cameras to make the detection of numerous objects in some specific areas. In most situations, using multiple cameras to make the multi-areas detection is more general. The number of the camera is considered as a key factor to make the multi-objects detection on multi-areas \cite{2016A}. In some detection situations, the multi-areas are small that can be contained in one whole visual scope of one camera. These multi-areas can be considered as the miro–multi-areas. In these situations, for the consideration of the resource consumption and implementation cost, one camera is more preferred to be accepted to implement the monitoring functions. Using one simple camera to make the individual detection of multiple sub-area of a large area is useful to make the individual area detection analysis. To achieve this goal, one general design method is to directly divide the areas of the visual video window. However, the methods based on  this idea are not free and easy to change or reset the configuration. Moreover, the setting process is also not visual and difficult to achieve. To address this issue, this paper has provided a novel drawing idea that can directly draw the detection area on the real-time video display screen, which will simplify the configuration process and decrease the time used to make the background code change.

\subsection{Camera Monitoring and Management}
As one of the successful implementations of computer vision, the camera monitoring  is widely used in security and vision monitoring and management. Using the camera to gain real-time video of the specific areas is useful for e-monitoring and management \cite{2015People}. In the special implementation situations, it is required to use less camera to monitor more than one specific sub-areas on the visible scope of one camera. It will be a large consumption of human energy if only monitoring these sub-area by human eyes. In the situations with higher monitoring requirements, the entering and leaving of the special monitored objects are also the items being monitored. In these situations, the technologies of electronic fencing will be useful to address the problems \cite{MalawskiFilip2021DVIS}. For the methods to implement electronic fencing, the convenience and real-time operation with the fast-response performances are weak but significant for the video monitoring and management. To address this issue, this paper has provided a novel method to directly manage the monitoring areas on the visible scope of one camera by directly drawing the monitored sub-areas on the real time video display, which is easy to reset and supports multi-areas synchronous configuration.

\section{Research Methodology}

\subsection{Simplify the YOLOv5 Project}
\begin{figure}
	\centering
	\includegraphics[width=0.4\textwidth]{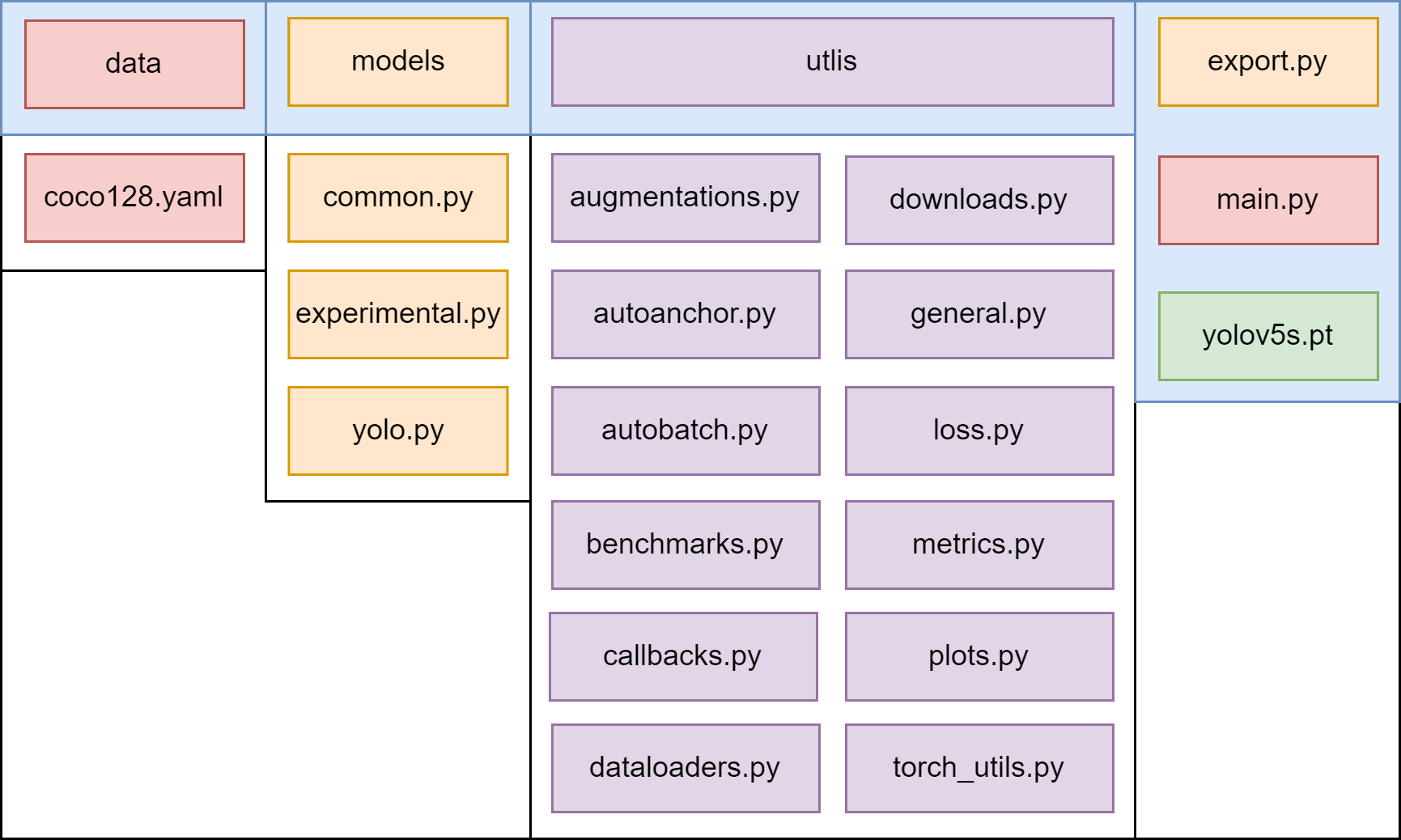}
	\caption{Simplified project structure.}
	\label{fig1}
\end{figure}

To improve the implementation efficiency of the project, the research of this paper has simplified the project structure of the YOLOv5. As shown in Figure \ref{fig1}, much of the previous folders and files have been deleted. Three related folders remain:$data$, $models$ and $utlis$. The key functional Python file of the research of this paper is file: $main.py$. Before the further design and development of this research, a reasonable and available simplification of the project structure will improve the efficiency of research and decrease the possibility of unexpected emergencies. In fact, the decrement of the unusable files of this previous YOLOv5,  will improve the direct implementation efficiency of this research.

\subsection{Display Video on Canvas}
\begin{figure}
	\centering
	\includegraphics[width=0.4\textwidth]{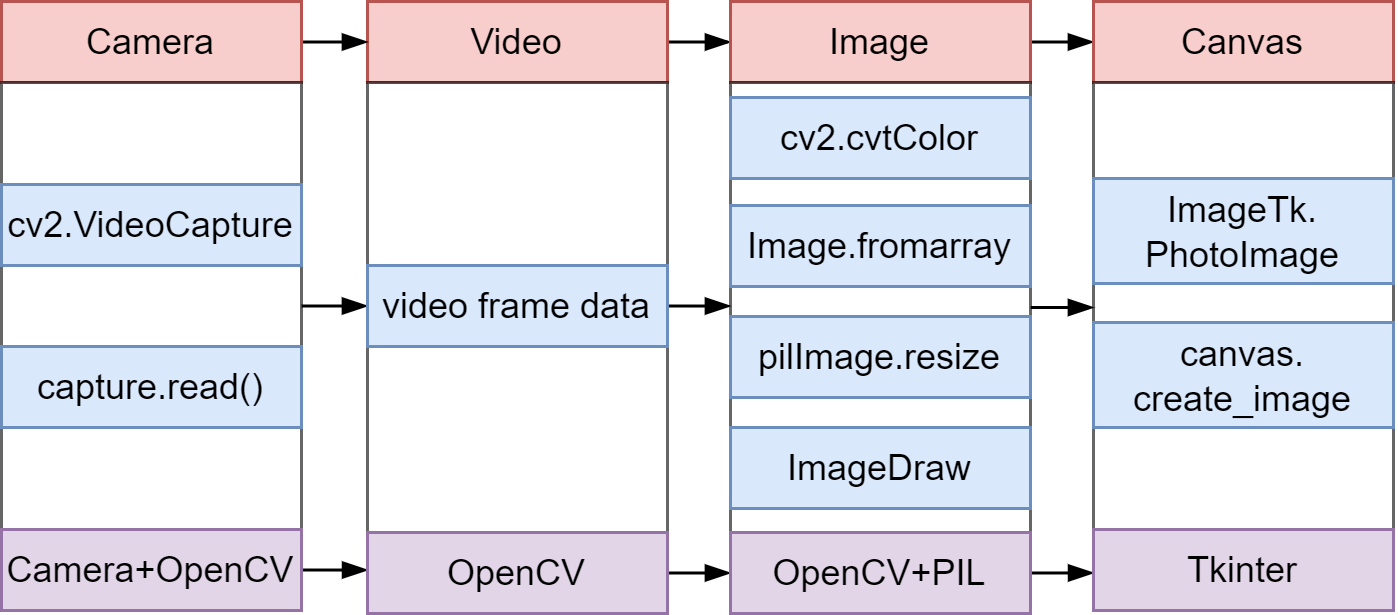}
	\caption{Drawing polylines detecting areas on video.}
	\label{fig2}
\end{figure}

Displaying the real-time video of the camera on a specific canvas is useful in terms of that using the canvas to design and generate the GUI will be easy and convenient. For the simple basic and pure Python developed program, the library Tkinter should be the first choice. The research of this paper, using the following design structure, as shown in Figure \ref{fig2}, implements the function of displaying the real-time video of the camera on the Tkinter canvas control, which can be embedded into the program developed by Tkinter. This process has realized that displaying a camera real-time video on a programmable Python project.

During the above process, as shown in Figure \ref{fig2}, firstly, the OpenCV will call the functions: $cv2.VideoCapture()$ and $capture.read()$ to open and get the video data, followed by the OpenCV gaining the video frame data from capturing the real time video of the camera. Subsequently, using the function component $cv2.cvtColor()$ to transform the format of the frame image array from $BGR$ to $RGBA$. Afterwards, using the $Image.fromarray()$ of PIL to gain the PIL  image object from the OpenCV array. For the better presentation of the video on the GUI, using the $pilImage.resize()$ function to resize the shape of the PIL image. Furthermore, if there is a need to draw the polylines, using the $ImageDraw()$ component of PIL to implement. Eventually, using the function $ImageTk.PhotoImage()$ to transform the PIL image to Tkinter image that can be used in the canvas control of Tkinter. Following this, using the $canvas.create_image$ to embed the real time frame image into the canvas. As the rapid transition of displaying pictures that are following the time series, the canvas image will display an performance of playing video.

The following is the key code of this research, which is used to implement the design principles mentioned above.

\begin{verbatim}
	cvimage = cv2.cvtColor(im0, 
	cv2.COLOR_BGR2RGBA)
	pilImage = Image.fromarray(cvimage)
	pilImage = pilImage. resize((image_width, 
	image_height), 
	Image.ANTIALIAS)
	dr_pilImage = ImageDraw.Draw(pilImage)
	tkImage = ImageTk.PhotoImage(image=pilImage) 
	root_window.children['!canvas']. 
	create_image(0, 0, anchor='nw', 
image=tkImage) 
	root_window.update() 
	root_window.after(1)
\end{verbatim}

\subsection{Drawing Polyline Areas on Video}

\begin{figure}
	\centering
	\includegraphics[width=0.4\textwidth]{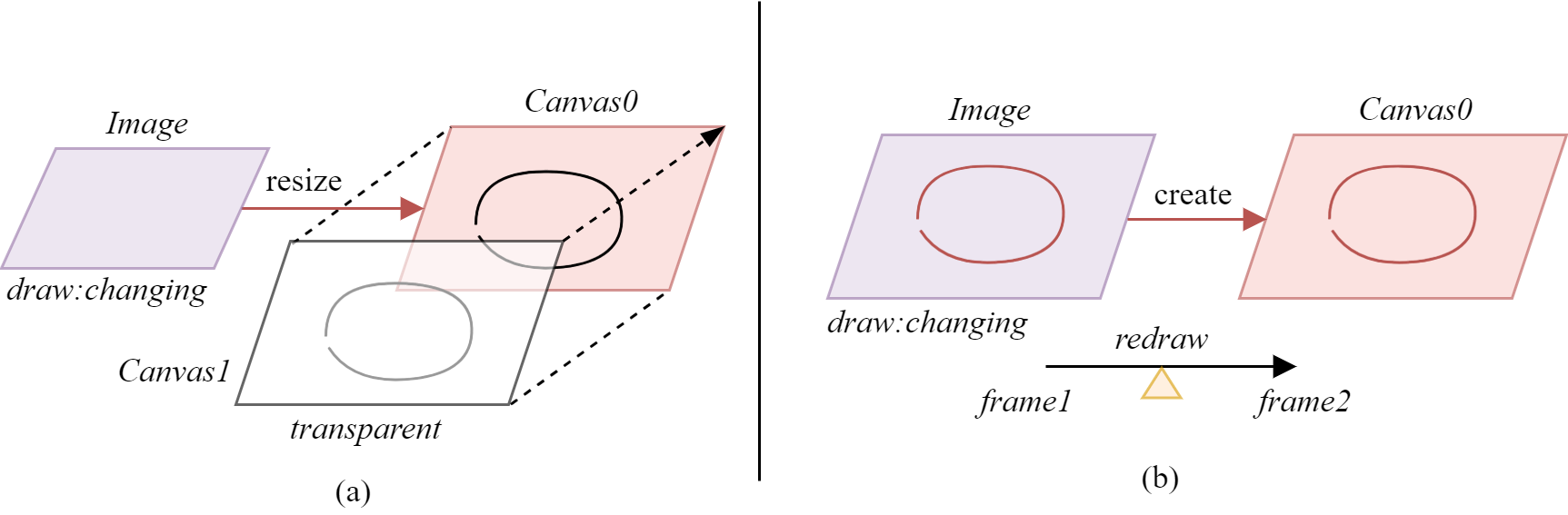}
	\caption{Drawing polyline detecting areas on video display.}
	\label{fig3}
\end{figure}

Drawing the free shape polylines to configure the limited detecting areas on the video display screen is the core novel idea of the research of this paper. After the research and experiments, for the normal considerations,  there are two main methods that seem available to implement this design. As shown in Figure \ref{fig3}, The first one, which is shown on the left a subfigure, is using a new canvas control $Canvas1$ of Tkinter to recover the previous canvas $Canvas0$ that was used to create the canvas images. In this approach, the background of the new canvas needs to configured as transparent. As the frame data of the video images changes, the drawing content of the $Canvas1$ will remain the same. But the way to make the background of the canvas control of Tkinter is unavailable for the current implementation or need to use the third-party methods. Therefore, method 2, which is shown on the right b subfigure of Figure \ref{fig3},  will be more available and was taken for this research. Compared with method 1, method 2 is more difficult in implementing the coding, and easier in the implementations of drawing and real time interactions. In method 2, the polylines will be redrawn between two continuous different frames, which means the performance of drawing is closer to real-time changing but needs more computed resource consumption in generating the final video edited images.

\begin{figure}
	\centering
	\includegraphics[width=0.4\textwidth]{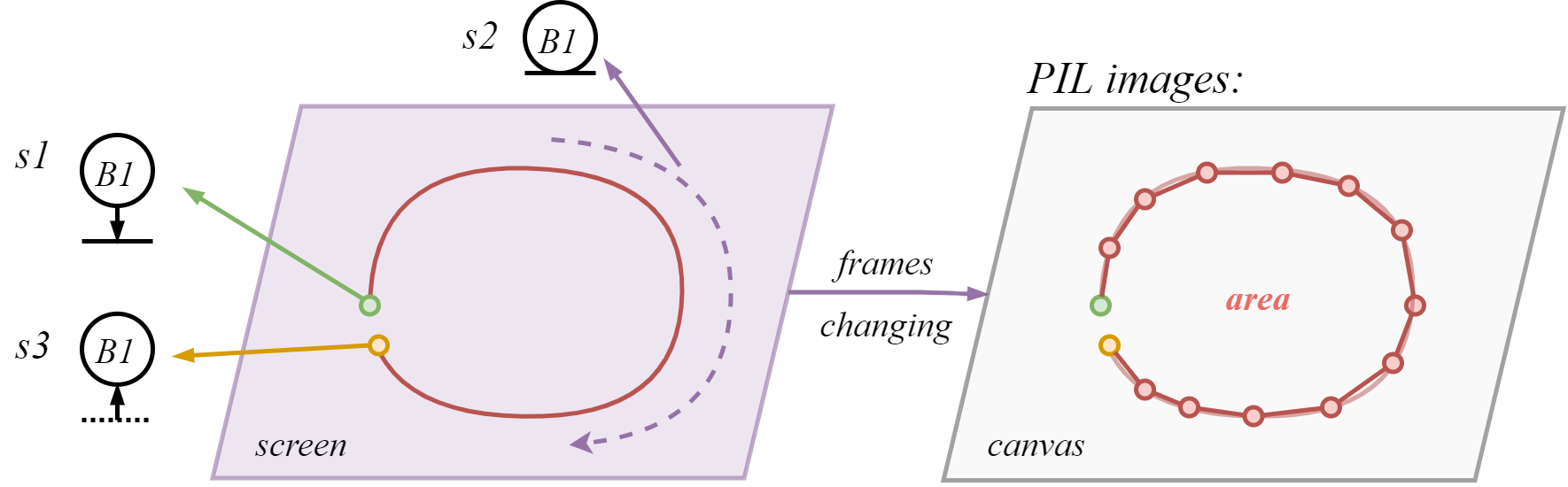}
	\caption{Interaction of mouse keys and showing the real time interactive performances on display. The coordinate data of the polylines will be gained in the background.}
	\label{fig4}
\end{figure}

In the implementation of the drawing on the display of the video of this paper, the main key is using the event handles of the mouse keys of Tkinter to gain the coordinate data of the drawing actions. As shown in Figure \ref{fig4}, $B1$ represents the left mouse button. When it comes to draw a detecting area, pressing the left mouse button to start the drawing procedure, which is called the $stage1$ and  represented by the label $s1$. Subsequently, continuously pressing the left mouse button to keep the running of the drawing procedure, moving the mouse to change the real time coordinates of the edge of the polylines area, which is called the $stage2$ and  represented by the label $s2$. Eventually, when the cursor comes to the terminal coordinate, releasing the left mouse button to finish the whole drawing procedure, which is called the $stage3$ and  represented by the label $s3$. For one drawing of one detecting area,  the drawing procedure can not be canceled during the drawing time. But the whole drawing of the display can be cleared out instantly by clicking the right mouse button. During the procedure of the drawing mentioned above, the recalling of the mouse button pressing and mouse cursor moving will gain the real time data of the moving track of the drawing polylines generated. As shown in Figure \ref{fig4}, as each frame changes to generate the PIL image, the new frame data will be generated followed by adding the drawing polylines data on the PIL image. In each different frame, one data of the new drawn mouse cursor edge point will be recorded. Finally, a detecting area, combined with many edge points which are located in different coordinates on the image is generated, with record of the coordinates data of the whole edge points. The data values of the edge points will be saved as a global variable and will not be changed as the transition of the different front-back two frames, unless the right mouse button is pressed to make a new drawing.

The following is the key code to implement that using the mouse button to make the interaction with Tkinter and recording the drawing data.

\begin{verbatim}
def b1_move(event):
    global draw_points_list
    global b1_moving
    draw_points_list.append((event.x, 
    event.y))
    if not b1_moving:
        b1_moving = True

def b3_press(event):
    global draw_points_list
    global draw_points_list_all
    draw_points_list.clear()
    draw_points_list_all.clear()

def b1_release(event):
    global b1_moving
    global draw_points_list
    global draw_points_list_all
    if b1_moving:
        draw_points_list_all.
        append(draw_points_list.copy())
        b1_moving = False
        draw_points_list.clear()
\end{verbatim}

\subsection{Detecting Key Points in Drawn Areas}

\begin{figure}
	\centering
	\includegraphics[width=0.4\textwidth]{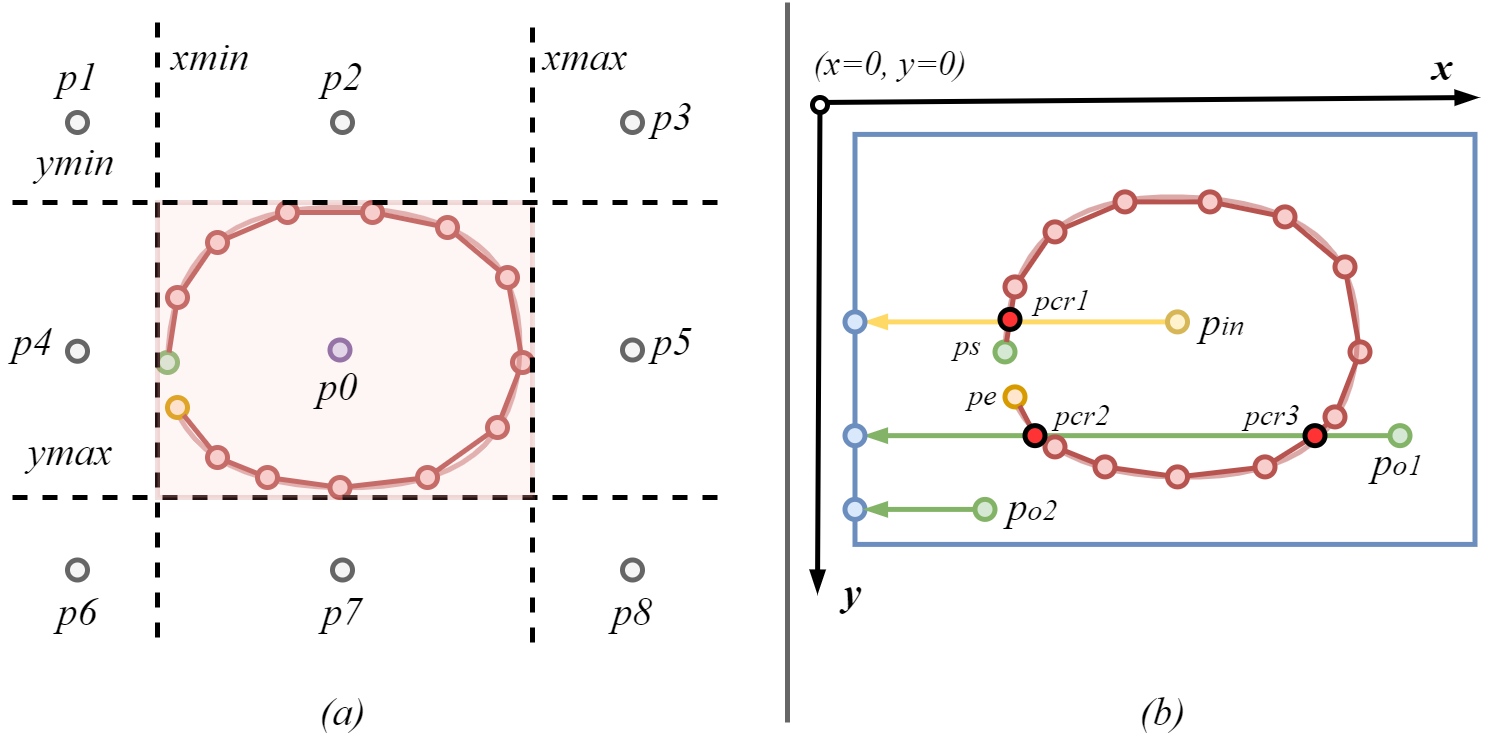}
	\caption{The processing of detecting the object key points in the visual screen image area. Subfigure $a$ shows that the whole single image 2D space area has been divided into nine sub-areas. Point $p0$ locates into the central area. Subfigure $b$ shows that point $p_{in}$ locates into the detecting area while points $p_{o1}$ and $p_{po2}$ do not.}
	\label{fig5}
\end{figure}

After gaining the data of the edge points that consist the polylines of the detecting area, and the data of the real time location coordinates of the detected people in the camera display visual area, the next step to detecting the human locating key points in the free drawn areas is to analyze whether the keypoint located into one or more of the drawn areas. As shown in Figure \ref{fig5}, when the analyzer of this research starts the detection of the key points of the human object, or other types of objects, the analyzer will analyze each key point’s $x$ and $y$ coordinates. The $x$ coordinate means the $x$ or width pixel distance of the key point from the left top corner of the image. The $y$ coordinate means the $y$ or height pixel distance of the key point from the left top corner of the image. If the coordinate of the one person key points do not locate into the central rectangle area of the nine divided sub-areas (e.g.: $p_{1}$, $p_{2}$, $p_{3}$, $p_{4}$, $p_{5}$, $p_{6}$, $p_{7}$, $p_{8}$), the analyzer will not implement the further crossing judgment and analysis for this key point. On the other hand, if the key points locating into the central rectangle area is detected, the analyzer of this research will run the judgment whether the key point locates into the drawn area or not.

Note that the shape of the draw area is not limited. Even there is a simple unclosed polyline that located on the left side of the key points, if the left horizontal (i.e.: follow the negative orientation of the $x$ axis ) ray of the key point intersects one the line segment of the polyline of the drawn area, the key point will also be identified as an inner point of the drawn area. Therefore, the method to draw a standard polyline that can consist of a detectable area is important. In most situations, a stable and closed polyline drawing is recommended. The judgment principle of the analyzer to judge whether a key point locates into a drawn area or not is using the ray casting principle. Based on this principle, the analyzer will cast a horizontal ray line to the left side edge of the image, followed by detecting the count of the points of intersection of the ray line and the segment lined of the polyline that built the drawn area. As shown in the subfigure $b$ of the Figure \ref{fig5}, point $P_{in}$ is located into the inner side of the drawn area with a count of the points of intersection is $1$(i.e.: odd number). While the points that are located in the outside of the drawn area, points $p_{o1}$ and $p_{o2}$, which have the counts of the points of intersection and the numbers are even numbers: 2 and 0. Here the number $0$ is also considered as an even number.  Due to the detecting principle of the analyzer of this research, to make a successful detection for the objects key points, the drawing of the polylines of the detecting area should be able to make the horizontal ray lines of the objects key points intersect themself. The polyline drawing of the detecting area can overlap and remain the individual detecting functions.

The following are the key codes to implement the design and development based on the principles and analysis mentioned above. The details can be found in the open source project link of this research that is mentioned in the section Introduction.

\begin{verbatim}
def get_area_max_min(area):
    …
def if_lines_cross(x11, y11, x12, y12, x21, 
    y21, x22, y22):
    …
def cross_detect(draw_points_list_all, 
    people_group, scale_w, 
scale_h):
    …
\end{verbatim}

\section{Experiments and Analyses}

\begin{figure}
	\centering
	\includegraphics[width=0.4\textwidth]{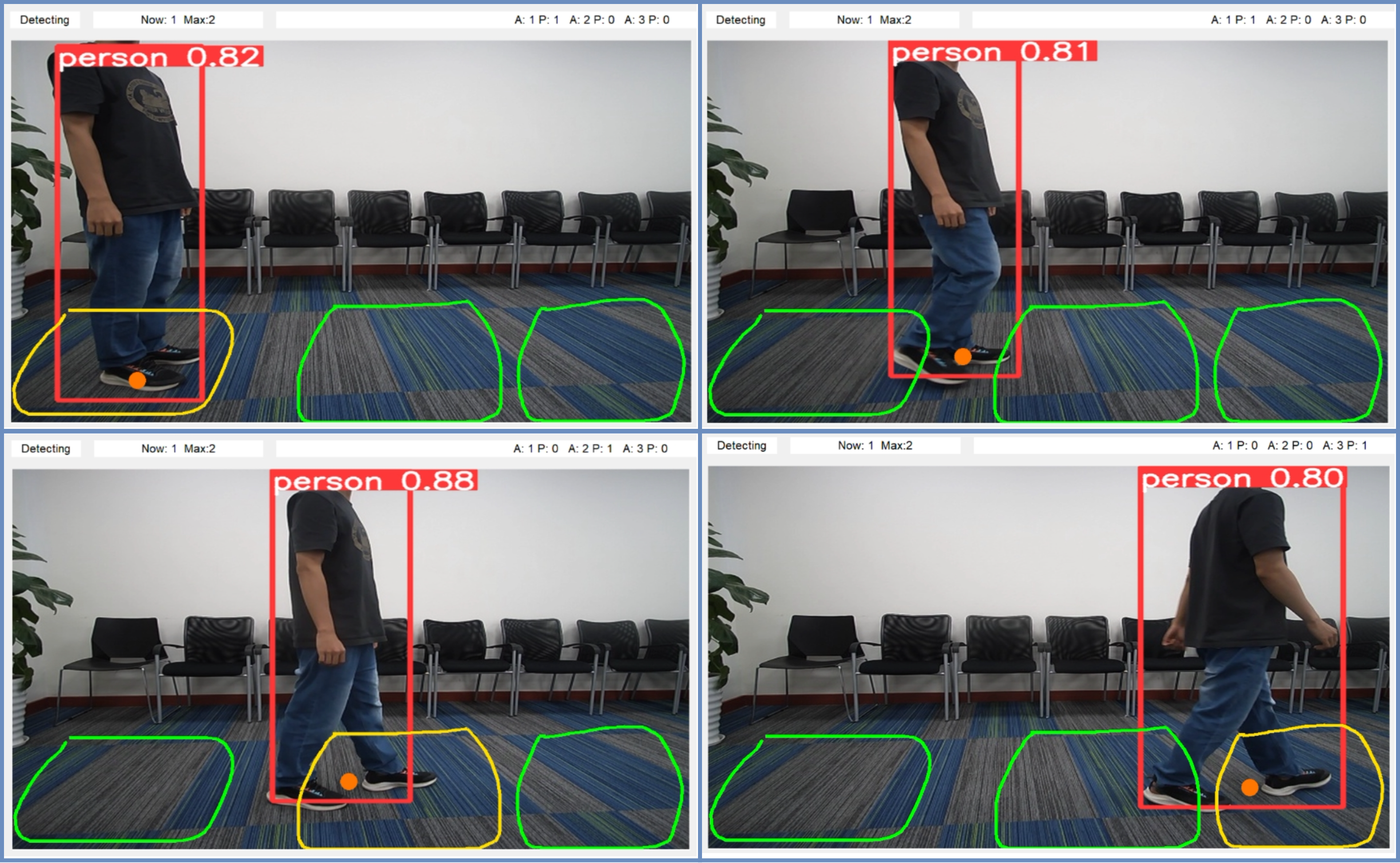}
	\caption{Using one camera to detect multiple sub-areas on the visual scope. The detected result is shown in the right top side of the display. The numbers of $A$ present the ID of detecting areas, and the numbers of $P$ present the count of the objects detected in the corresponding detecting sub-areas.}
	\label{fig6}
\end{figure}

The research of this paper using a real time running camera to test the implementation of the design and algorithm principles of the paper. As shown in Figure \ref{fig6}, to make the experiment more understandable visually, the key point of the human object was drawn as a brown point. When the object has entered the detecting areas, the color of the outlines of the areas will change from green to yellow. When the detected object leaves the detecting area or enters the gap between two detecting areas and does not locate into any detecting areas, the outline color of the previous detecting areas will return to the previous green. The detection for the object is the detection for key points of the objects. After the single and multiple experiments, the design and algorithms principles of this research are tested as useful and available. The processing efficiency of the detecting procedure is high and causes little influence on the recognition model running, due to its simplified coding and rapid efficiency.

\begin{figure}
	\centering
	\includegraphics[width=0.4\textwidth]{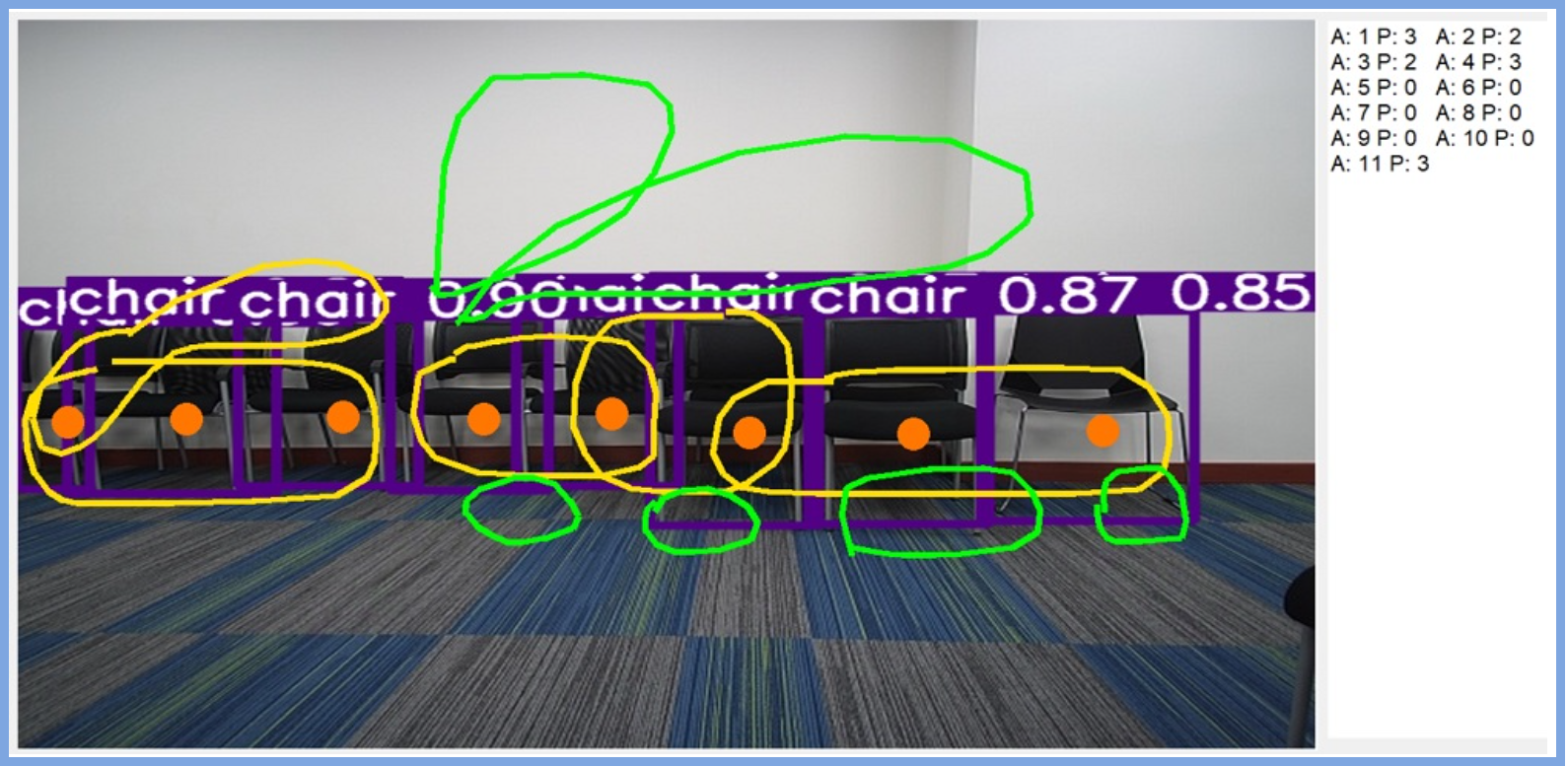}
	\caption{The overlapping detecting sub-areas in one camera visual scope. The detected result is shown on the right side.}
	\label{fig7}
\end{figure}

The drawing function of the research of this paper is able to be overlapped. As shown in the Figure \ref{fig7}, the individual detecting areas can be overlapped and only detect the object that locates into the inner area of the detecting area. This characteristic will be useful when it comes to the situations of crossing multiple areas detection. Eventually, the free drawing detection function can be also used to detect other classes of objects except human objects, for instance, detecting the chairs. The detecting modes and detection accuracy and efficiency is related to the specific recognition model. But this free drawing detecting implementation will not influence the efficiency of the recognition model too much. By the test experiments, it can be gained that this multiple detecting design and principle is available and useful.

\section{Conclusion}

The research of this paper has provided a reasonable and available design principle and the implementation of the whole development idea and analyses of the corresponding open source code. The design and implementation idea and code of this research can be used to implement the multiple detection of different sub-area on one camera visual scope. The implementation methods have been proved to be effective. The free drawing of the outlines of the detecting areas make it easy to implement the customized detection for the multiple inner areas with one camera, which will reduce the consumption of the hardwares, such as monitor cameras and the corresponding manager instruments. The implementation idea and program code can be extended to numerous monitoring and detecting situations, which means the implementation principle is useful and meaningful.

\bibliographystyle{IEEEtran} 
\bibliography{ref}



\end{document}